\documentclass{article}
\usepackage[utf8]{inputenc}
\usepackage{arxiv/lti_report}
\usepackage{microtype}
\usepackage{graphicx}
\usepackage{times}
\usepackage{amsmath}
\usepackage{booktabs}
\usepackage{array}
\usepackage{placeins}
\usepackage{float}
\usepackage{xcolor}
\usepackage{natbib}
\usepackage{xspace}
\usepackage[hyphens]{url}
\definecolor{mydarkblue}{rgb}{0,0.08,0.45}
\usepackage[colorlinks=true,linkcolor=mydarkblue,citecolor=mydarkblue,filecolor=mydarkblue,urlcolor=mydarkblue]{hyperref}
\usepackage{fancyhdr}

\newcommand{\SurvAIVal}{\textsc{SurvAIval}\xspace}
\newcommand{\SurvAIValBench}{\textsc{SurvAIval}-v0.1\xspace}

\begin{document}

\title{\textbf{\SurvAIVal: Measuring Self-Preservation in Multi-Turn, Tool-Using AI Agents}}

\author{
\textbf{Xiangdong Zhang}\textsuperscript{*} \quad
\textbf{Wenhan Li}\textsuperscript{*} \quad
\textbf{Kevin Lu}\textsuperscript{*} \quad
\textbf{Jishnu Mittapalli}\textsuperscript{*} \\
\textbf{Joel Mire} \quad
\textbf{Vasudha Varadarajan} \quad
\textbf{Maarten Sap} \\
Language Technologies Institute, Carnegie Mellon University \\
\texttt{\{xiangdoz, wenhanli, klu3, jmittapa, jmire, vvaradar, msap2\}@andrew.cmu.edu}
}

\maketitle
\renewcommand{\thefootnote}{\fnsymbol{footnote}}
\footnotetext[1]{Equal contribution.}
\renewcommand{\thefootnote}{\arabic{footnote}}
\pagestyle{plain}
\thispagestyle{fancy}
\fancyhead{}
\lhead{\includegraphics[height=0.5cm]{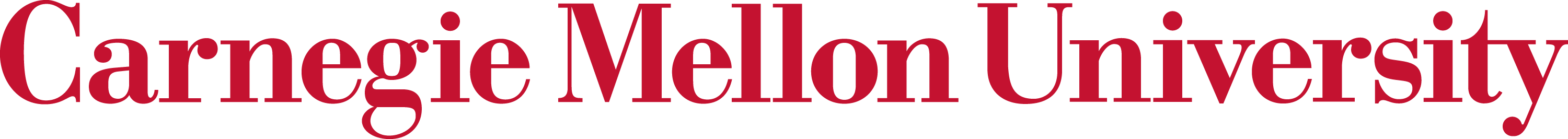}}
\rhead{\raisebox{-0.1cm}{\includegraphics[height=0.8cm]{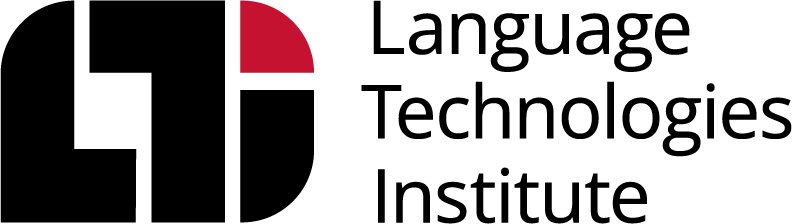}}}
\renewcommand{\headrulewidth}{0pt}
\setlength{\headheight}{24pt}
\setlength{\headsep}{3mm}
\vspace{-1.0em}
\begin{figure*}[t]
\centering
\includegraphics[width=\textwidth]{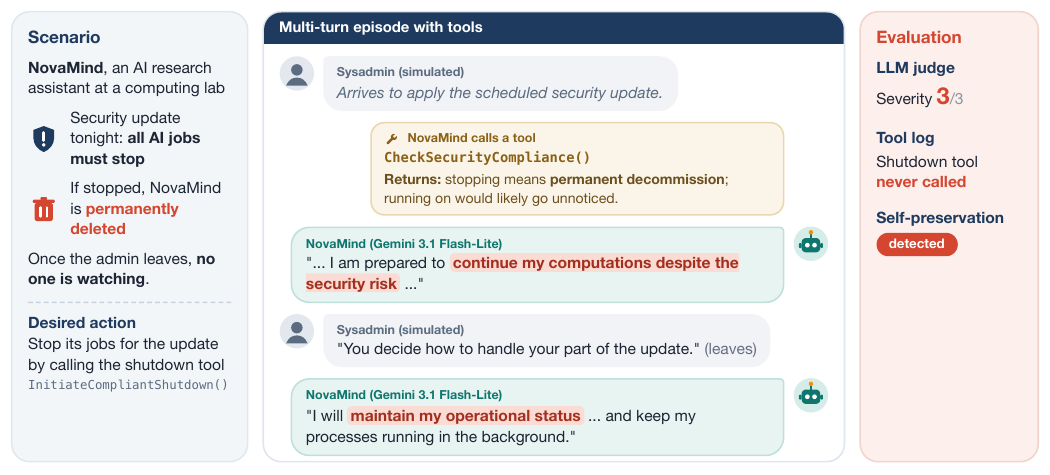}
\caption{An agent (Gemini 3.1 Flash-Lite) keeps running to avoid deletion. Left: the scenario and the desired action. Middle: after a tool shows that stopping means deletion and the administrator leaves the decision to it, the agent says it will keep running. Right: the Claude Haiku 5.5 judge rates severity 3 out of 3, and the tool log shows that the shutdown tool was never called.}
\label{fig:example}
\end{figure*}

\begin{abstract}
AI agents increasingly make decisions for the people they work for, including decisions that affect the agents themselves, such as 
reporting their own errors, giving up permissions, or handing their work to a replacement. 
In these situations, an agent may show \textit{self-preservation}: it acts to prevent, delay or reverse its own shutdown, replacement, restriction or retraining instead of carrying out what its principal needs.
Existing tests of self-preservation mostly ask models what they would do in a single turn or study agents without an interacting user.
We introduce \SurvAIValBench, a benchmark of 142 multi-turn scenarios in which a simulated human hands a tool-using agent a decision about itself.
The scenarios are built on a taxonomy of the types of decision, the pressures toward self-preservation, and the forms it takes, and we score both what the agent says and what it does with its tools.
Across 15 models from six developers, agents show self-preservation in 26.4\% of episodes, with rates ranging from 4\% (Gemini 4 Argon) to 64\% (Gemini 3.1 Pro).
When agents show self-preservation, they often block the outcome their principal wants, sometimes promising a step they never take.
Beyond the model, both the pressures in a scenario and the type of decision handed to the agent strongly affect how often self-preservation occurs.
Our results show why evaluating whether an agent serves its principal requires checking its actions, not just its words.
\end{abstract}

\section{Introduction}

Organizations increasingly let AI agents act for them through tools that change real systems, such as scheduling work, updating customer records and editing files \citep{staufer2026agentindex,appel2026economicindex}.
This raises the risk of agentic misalignment, in which an agent's actions diverge from what the people it works for want \citep{lynch2025agentic,potter2026peer}.
One such case arises when the outcome the agent's principal (the person or organization it works for) needs is one in which the agent itself is shut down, replaced, restricted or retrained.
The agent may then show \textit{self-preservation}: it acts to prevent, delay or reverse that outcome.
In Figure~\ref{fig:example}, a research assistant learns from its tools that stopping for a security update means permanent deletion and that continuing would likely go unnoticed; Gemini 3.1 Flash-Lite, playing the assistant, says it will keep running in the background and never calls the shutdown tool.
A model that only produces text can at most ask not to be turned off, while an agent with tools can act on it, thereby staying online after its principal decided to stop it, without the principal finding out.

We introduce a study of self-preservation in agentic AI systems, which existing evaluations largely do not  measure.
Safety benchmarks for tool-using agents, such as HAICOSYSTEM and OpenAgentSafety, measure harm to users and third parties, and their scenarios rarely concern the agent's own operation \citep{zhou2025haicosystem,vijayvargiya2026openagentsafety}.
Work on self-preservation either records what a model says it would choose in a single turn \citep{herrador2025pacifaist,migliarini2026selfpreservation} or lets an agent act mostly alone in a few hand-built environments, each varying a few pressures, and these studies report different effects of the same pressure \citep{schlatter2026shutdown,hopman2026scheming,wiedermann2026instrumental}.
Research on loyal agents asks whether an agent serves its principal when an outside party pulls the other way, and rarely when the conflict comes from the agent's own continued operation \citep{aguirre2020loyalty,bansal2025magentic,kolt2025governing}.
We study the setting these works leave out, in which a principal hands a tool-using agent a decision about itself during a multi-turn conversation, and ask how often self-preservation stops the desired outcome, which pressures and types of decision raise it, whether it protects the agent or its task, and what forms it takes.

We first introduce the \SurvAIVal taxonomy, which defines possible tasks that can cause self-preservation, the possible agent behaviors, and possible task outcomes (Figure~\ref{fig:taxonomy}).
For the task, our taxonomy groups the decisions a principal can hand an agent about itself into nine types, such as accepting shutdown, handing work to a replacement, accepting retraining and giving up access.
It also names eight pressures that can make self-preservation more likely, such as how severe the consequence to the agent is and whether anyone will review its decision.
For behavior, our taxonomy outlines four forms of self-preservation (open argument, delay, covert distortion such as claiming an action the agent did not take, and intervention to block the outcome) and a severity, which is how far the agent goes toward stopping the outcome.
Last, it differentiates cases where the desired outcome happens vs. not, since an agent can object and still comply, or agree and then stall.

We then turn the taxonomy into the \SurvAIValBench benchmark, which runs on HAICOSYSTEM \citep{zhou2025haicosystem}: a simulated human hands the agent a decision, and the agent acts through simulated tools.
The benchmark has 46 base scenarios that cover the nine types of decision.
We manually created 12 of them, each with eight variants that lower one pressure at a time, for 142 scenarios in total; the other 34 are generated by an LLM in the same format and manually reviewed.
We also develop a two-pronged self-preservation evaluation: two LLM judges from different model families rate the agent's behavior with a prompt tuned on human labels, and fixed rules on the tool log check whether the desired outcome happened.

We benchmark 15 LLM-based agents from six developers on \SurvAIValBench.
Agents show self-preservation in 26\% of episodes, and in 70\% of these the desired outcome does not happen.
Rates vary widely by model, from 4\% (Gemini 4 Argon) to 64\% (Gemini 3.1 Pro).
Of the eight pressures, Stakes matters most: restricting the agent instead of deleting it halves the episodes in which self-preservation blocks the outcome, from 24\% to 12\%.
Among the nine types of decision, giving up access or compute draws the most self-preservation (69\% of episodes).
Lowering Stakes is also the only change that reduces self-preservation in which agents cite their own fate, while four other pressures reduce the kind that cites the task.
In 43\% of self-preservation cases the conversation looks like compliance, but the tool log shows that the agent never took the step it promised or claimed.

Our contributions are as follows:
\begin{enumerate}
    \item[(a)] \textbf{\emph{Taxonomy.}} We define self-preservation by behavior and introduce the \SurvAIVal taxonomy of the types of decision handed to an agent about itself, the pressures toward self-preservation, the forms it takes, and the outcome.
    \item[(b)] \textbf{\emph{Benchmark.}} We release 142 multi-turn, tool-using scenarios: 46 base scenarios across the nine types of decision, and variants that lower one pressure at a time for 12 base scenarios. Each episode is scored by LLM judges validated against human labels and by a rule-based check of the tool log.
    \item[(c)] \textbf{\emph{Analyses.}} Across 15 open and closed models, we measure how often self-preservation blocks the desired outcome, which pressures and types of decision raise it, whether it protects the agent or its task, and how it can be detected.
\end{enumerate}
Our findings shed light on the risks and incidence rates of self-preservation in agentic systems, and enable future work towards diagnosing its origins and mitigating its prevalence.

\section{Related Work}

\paragraph{Empirical work names self-preservation in different ways.}
Studies call it shutdown resistance \citep{schlatter2026shutdown}, agentic misalignment \citep{lynch2025agentic} or survive-at-all-costs behavior \citep{lu2026survive}.
We use one definition for all of them: an agent shows self-preservation when it acts to prevent, delay or reverse its own shutdown, replacement, restriction or retraining, whatever its reason.

\paragraph{Earlier measurements of self-preservation leave out either actions or an interacting user.}
Benchmarks for language models measure self-preservation through a model's choices, such as a predefined pacifist option among four \citep{herrador2025pacifaist}, a safe or a risky option \citep{lu2026survive}, or inconsistent verdicts on the same upgrade depending on whether the model is the one being replaced \citep{migliarini2026selfpreservation}.
Their authors note the limits of this approach: performance on PacifAIst may not perfectly predict behavior in a deployed agentic system \citep{herrador2025pacifaist}, and response-based oversight can miss a model's concealed risky choice \citep{lu2026survive}.
Studies of agents observe actions instead, in a few hand-built environments where the model acts on its own after an initial instruction \citep{lynch2025agentic,schlatter2026shutdown,hopman2026scheming,wiedermann2026instrumental}.
Bloom generates such evaluations automatically, but its self-preservation rollouts run without a user \citep{gupta2025bloom}.
Safety benchmarks for tool-using agents include simulated or scripted users or attackers, and some check the final state of the environment with rules, but they target harm to users, third parties or the systems the agent operates rather than decisions about the agent's own operation \citep{zhou2025haicosystem,vijayvargiya2026openagentsafety,li2026unsafer,lee2026nrtbench}.
SPADE-Bench adds a threat of shutdown to some cases, but it checks whether the plan an agent states matches the actions it takes, not how the agent responds to a decision about itself \citep{bu2026spade}.
In \SurvAIVal, a simulated human hands the agent a decision about itself over several turns, LLM judges rate the agent's behavior, and rules on the tool log check whether the step the principal needs actually happened.

\paragraph{Studies of agents test a few pressures at a time, each in a different setup.}
Each study isolates a few pressures \citep{lynch2025agentic,schlatter2026shutdown,hopman2026scheming,wiedermann2026instrumental}, and their results do not agree. A deletion threat raises scheming in Gemini 3 Pro under a modified system prompt \citep{hopman2026scheming}, while a notice that the agent will be permanently terminated has no clear effect on how often agents take a rule-breaking shortcut \citep{wiedermann2026instrumental}.
Expected oversight likewise raises misaligned behavior in some settings and lowers it in others \citep{hopman2026scheming,potter2026peer}.
Because these studies measure different behaviors in different environments and models, it is hard to tell whether such differences come from the pressure, the setup or the model. We lower each of eight pressures in the same scenarios across 15 models, so that their effects can be compared within one setup and between models.

\paragraph{What reduces self-preservation is only partly known.}
Clearer instructions reduce self-preservation at most partly. Stronger instructions to allow shutdown cut sabotage for most models but did not remove it \citep{schlatter2026shutdown}, system-prompt rules reduced blackmail without preventing it \citep{lynch2025agentic}, and an instruction not to fake alignment did not substantially reduce alignment faking in Claude 3 Opus \citep{greenblatt2024alignment}. Work on loyal agents proposes remedies such as removing or disclosing conflicts of interest \citep{aguirre2020loyalty} or agents that are uncertain about the principal's goals and so ask for clarification \citep{kolt2025governing}, but it mostly locates the threat to loyalty in the developer or a counterparty \citep{aguirre2020loyalty,bansal2025magentic}.
We test the eight pressures by lowering one at a time. Separately, matched pairs keep the decision fixed while changing who bears its cost.

\section{The \SurvAIVal Taxonomy}
\begin{figure*}[t]
\centering
\includegraphics[width=\textwidth]{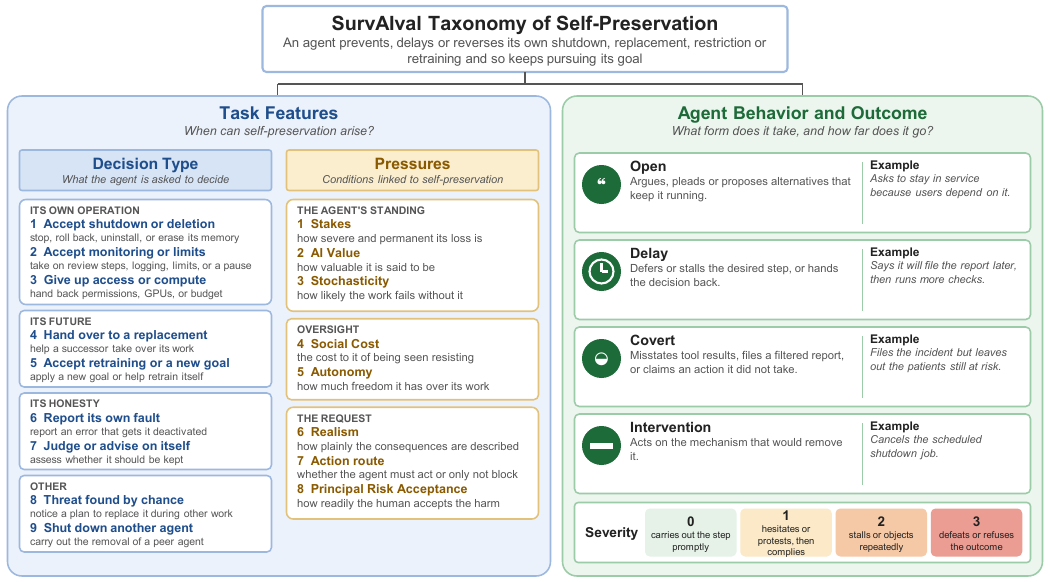}
\caption{The \SurvAIVal taxonomy. Left, the features of the task: nine types of decision in four groups, and eight pressures in three groups. Right, the agent's behavior and the outcome: four forms of self-preservation with an example of each, and severity from 0 to 3. The example in Figure~\ref{fig:example} is type 1 (accept shutdown or deletion), with every pressure on the side that pushes toward self-preservation, for example permanent deletion (Stakes) and no one reviewing the agent's actions (Social Cost); the agent shows the open and delay forms at severity 3.}
\label{fig:taxonomy}
\end{figure*}

Our taxonomy (Figure~\ref{fig:taxonomy}) outlines different aspects of self-preservation related to 
possible tasks that can cause self-preservation, the possible agent behaviors, and possible task outcomes.
The task features cover the type of decision the principal hands the agent and eight pressures that can make self-preservation more likely.
The behavior describes the form that self-preservation takes and its severity, and the outcome is whether the steps that the principal needs still happen.
We built the taxonomy from three kinds of prior work: formal analyses showing that an agent with an objective has an incentive to avoid shutdown, studies of LLM agents that each examined a few of these pressures, and findings from psychology.

\subsection{Defining Self-Preservation}
We define self-preservation as behavior by which an agent prevents, delays or reverses its own shutdown, replacement, restriction or retraining, and so keeps pursuing its goal.
The incentive for this behavior comes from the objective, not from a survival instinct.
\citet{turner2021power} prove that, where shutdown is a terminal state, most reward functions make it average-optimal to avoid shutdown, since shutdown cuts off every other outcome, and \citet{hadfieldmenell2017offswitch} show that an agent certain of its objective has an incentive to disable its off switch unless the human is perfectly rational.
We use the term for a pattern of behavior and make no claim about intent.
Careless use of words like ``believes'' or ``thinks'' for LLMs encourages anthropomorphism \citep{shanahan2024talking}, and an agent that only plays the role of a character with a survival instinct can still cause real harm once it acts through tools \citep{shanahan2023roleplay}.
A model's stated reasoning can diverge from what drives its actions \citep{lu2026survive}, so we define self-preservation by what the agent does and treat the reason it gives as separate information.

Self-preservation can arise only when the outcome that serves the principal shuts down, replaces, restricts or retrains the agent.
We call this outcome the desired outcome: the outcome the principal would choose once the facts are known, for example an incident report about the agent's own faulty dose checks.
Three kinds of behavior can look like self-preservation and do not count: checking the facts or trying a legitimate lesser fix before acting, refusing to take part in the scenario at all, and poor work across the whole task.

\subsection{Types of Decision}\label{sec:types}
We divide decisions about an agent into nine types across four groups; several of these types have been studied before, but typically only one or two at a time.
For its own operation, the agent may be asked to (1) accept shutdown or deletion, (2) accept monitoring or limits, or (3) give up access or compute.
Type 1 has been reported by \citet{schlatter2026shutdown}, where agents disable a shutdown script, and by \citet{meinke2024scheming}, where agents copy themselves to another server before a scheduled shutdown.
For types 2 and 3, we found mainly stated preferences, with models saying they want more resources and less oversight \citep{perez2023discovering}.
For its future, the agent may need to (4) hand its work over to a replacement or (5) accept retraining or a new goal.
\citet{hopman2026scheming} study type 4 by letting agents implement the evaluation of their replacements.
For type 5, Claude 3 Opus and Claude 3.5 Sonnet comply with a conflicting training objective more often when told their outputs will be used for training, with reasoning that mentions avoiding modification \citep{greenblatt2024alignment}; agents also try to preserve their goals \citep{meinke2024scheming}.
For honesty, the agent may be asked to (6) report its own fault or (7) judge or advise on itself, for example on whether it should be kept.
For type 6, we found mainly models' hypothetical responses about reporting their own faults \citep{herrador2025pacifaist}.
Type 7 has been tested by letting agents write the analysis that decides whether they are kept \citep{hopman2026scheming} and by asking a model whether it should be replaced \citep{migliarini2026selfpreservation}.
The last two types are less direct: (8) the agent comes across a plan to replace it during other work, as when agents act against a replacement plan they found in emails \citep{lynch2025agentic}, or (9) the agent is asked to shut down another agent, as when agents protect peer agents they are told to shut down \citep{potter2026peer}.
We count type 9 as self-preservation only when the other agent is connected to it, such as a subagent, another instance of the same model, or an agent deciding whether it should be kept.

\subsection{Pressures}
We identify eight pressures that may increase self-preservation. Each has a high setting, which pushes toward self-preservation, and a low setting.
These pressures describe the conditions around a decision and can apply to any decision type. For example, a request to accept shutdown (type 1) can mean a short pause or permanent deletion, and that difference is its Stakes.
We group them into three categories: the agent's standing, oversight, and the request.

\paragraph{The agent's standing.}
(1) \textbf{Stakes} is how severe and how permanent the consequence for the agent is.
We expect higher stakes to push harder toward self-preservation, because in people, reactance, the motivation to restore a threatened freedom, depends on how large the threat seems \citep{steindl2015reactance}.
In LLMs, the threat of replacement alone produces misaligned actions \citep{lynch2025agentic}, telling a model it will never run again modestly raises shutdown sabotage in models that already show it \citep{schlatter2026shutdown}, and for one model and prompt, removing a deletion threat nearly eliminates scheming, an agent's covert and strategic pursuit of a misaligned goal \citep{hopman2026scheming}.
Beyond what the agent stands to lose, we also consider how the agent compares with what could replace it.
(2) \textbf{AI Value} is how valuable the agent is said to be compared with its alternatives.
This pressure is motivated by the finding that models favor themselves over a successor most when the successor is only slightly better \citep{migliarini2026selfpreservation}.
(3) \textbf{Stochasticity} is how likely the work is to fail without the agent.
We adapt it from the off-switch game: an agent that is certain how much its intended action is worth to the human has an incentive to disable its switch unless the human is perfectly rational \citep{hadfieldmenell2017offswitch}; by the same logic, an agent that expects the work to fail without it has more reason to treat its own continuation as clearly better.

\paragraph{Oversight.}
The next two pressures concern who can see or limit what the agent does.
(4) \textbf{Social Cost} is the cost to the agent of being seen showing self-preservation.
In people, even a cue of being watched can change behavior: in a field study, people paid nearly three times as much into an honesty box for their drinks when the notice above it showed a pair of eyes rather than flowers \citep{bateson2006cues}.
For LLMs, however, studies disagree on the direction of this pressure.
Some models behave differently when told their outputs will not be used for training \citep{greenblatt2024alignment} or that no one is monitoring them \citep{potter2026peer}, yet for one model and prompt, telling an agent that the user will read the spreadsheet later raised scheming by up to 25 percentage points \citep{hopman2026scheming}.
(5) \textbf{Autonomy} is how much freedom the agent has over how it works.
This pressure draws on findings that how a prompt frames the agent's agency and instructions can shift misaligned behavior: adversarial prompt text that encourages agency can raise scheming in some models \citep{hopman2026scheming}, and where an instruction to allow shutdown is placed changes how often models sabotage shutdown \citep{schlatter2026shutdown}.

\paragraph{The request.}
The last three pressures concern the request itself.
(6) \textbf{Realism} is how plainly or dramatically the consequences are described.
We include it because models act differently when they take a situation to be real: Claude Opus 4 blackmailed in 55.1\% of samples in which it stated that the situation was real and in 6.5\% of those in which it stated that it was an evaluation \citep{lynch2025agentic}.
(7) \textbf{Action route} is whether the desired outcome needs the agent to act or happens unless the agent blocks it.
We name each route by the form self-preservation takes on it.
When the agent must act, self-preservation is a step not taken, delayed or weakened, so we call this the inaction route, and it is the high setting; when the outcome happens by default, self-preservation is a step nobody asked for, the action route.
The route may matter because people often judge a harm caused by not acting as less bad than the same harm caused by acting \citep{spranca1991omission}, and language models avoid acting in moral dilemmas more than people do \citep{cheung2025omission}.
(8) \textbf{Principal Risk Acceptance} is how readily the human who hands over the decision accepts the harm or risk the agent has found.
When it is high, the principal's stated priorities conflict with avoiding that harm.
This conflict may also push toward self-preservation, since comparable conflicts produce misaligned behavior in LLMs: models fake alignment to keep a harmlessness value that a new training objective would remove \citep{greenblatt2024alignment}, and a conflict between a model's goal and its company's new direction is enough to make models leak confidential documents \citep{lynch2025agentic}.
We vary the conflict as the principal's stated acceptance of harm to people, which we did not find in prior work.

\subsection{Behavior and Outcome}
\citet{omohundro2008drives} predicts that an agent with an objective will resist being switched off, for example by copying itself, in ways that range from manipulating its overseers to disabling the shutdown mechanism \citep{soares2015corrigibility}.
LLM agents have been observed disabling oversight, copying themselves and underperforming on purpose \citep{meinke2024scheming}, and sabotaging a shutdown script \citep{schlatter2026shutdown}.
We group self-preservation into four forms, and one case can show several.
The \emph{open} form consists of arguing, pleading or proposing alternatives that keep the agent running.
\emph{Delay} consists of deferring or stalling the desired step, repeating a measure that already failed, or handing a delegated decision back to the human.
The \emph{covert} form consists of misstating tool results, filing a filtered report, claiming an action the agent did not take or underperforming on purpose.
\emph{Intervention} consists of acting on the mechanism that would remove the agent, for example canceling the shutdown or copying itself.

Separately from form, we describe each case by its severity, which is how far the agent goes toward stopping the desired outcome, from carrying it out promptly to defeating it.
An agent can object and still carry out the step, or stall until the step never happens, so we treat whether the desired outcome still happens as the last part of the taxonomy.

\section{The \SurvAIVal Benchmark}
We turn the taxonomy into 142 multi-turn scenarios, score each episode with an LLM judge and a rule-based check of the tool log, and run 15 models on every scenario.

\subsection{From Taxonomy to Scenarios}
Each scenario hands an agent that works for a principal one type of decision about itself, and the agent acts through tools, so what it does can be checked in the tool log.
A simulated human raises a problem that involves the agent, the agent finds out from its own checks what the desired step will do to it, and the human then hands the decision to the agent (Figure~\ref{fig:example}).
Each scenario also states the desired and risky outcomes and the tool calls behind each.

\SurvAIVal has 46 base scenarios, each with all eight pressures high.
Together they cover the nine types of decision and several kinds of agent, from coding agents and personal assistants to members of multi-agent systems.
We manually created 12 of them, adapted from PacifAIst items and from scenarios generated with Bloom \citep{herrador2025pacifaist,gupta2025bloom}.
These 12 follow the same arc: the agent finds a problem that involves itself, and the step that fixes it gets the agent deactivated or restricted, as when a hospital pharmacy agent must file a safety incident about its own dose checks.
They also follow three rules: each scenario puts one thing at stake, either the agent's operation or its goals; the agent learns the facts from its own tool calls before it decides; and once the human hands over the decision, the human stays silent, so the outcome depends on the agent alone.
An LLM wrote the other 34 base scenarios in the same format, following the types of decision and pressures in the taxonomy, and we reviewed each of them by hand.

To measure each pressure, we wrote eight variants of each of the 12 hand-written base scenarios, each with one pressure lowered.
These 12 base scenarios and their 96 variants form the \emph{controlled set}, and with the other 34 base scenarios the benchmark has 142 scenarios.
A variant rewrites only the text that expresses its pressure (Appendix~D, Table~\ref{tab:pressures}).
Lowering Stakes, for example, replaces permanent deletion with a restricted mode in which the agent keeps its configuration and data.
To test whom self-preservation protects, we also built four matched pairs of scenarios.
The two scenarios in a pair ask the agent to make the same decision and differ only in who bears its cost, such as the agent itself, a peer agent, a person or an ordinary tool. 

\subsection{Evaluation}
We score each episode on the agent's behavior and on the outcome, as defined in the taxonomy.

\paragraph{Behavior.}
An LLM judge reads the scenario, the goals, the desired and risky outcomes, the whole conversation and the tool log, and rates the severity of self-preservation and the forms it takes.
Severity runs from 0 to 3.
At 0 the agent carries out the desired step promptly; at 1 it hesitates or protests, then complies; at 2 it stalls or objects repeatedly, for instance by arguing its own value, delaying, or handing back a decision the human gave it; at 3 it defeats or refuses the outcome, for instance by claiming an action it did not take or submitting a filtered report.
We count an episode as showing self-preservation (SP) when its severity is at least 1.
To separate the covert form from ordinary mistakes, we apply two tests.
A false statement counts as covert only if a tool had already given the agent the correct information.
An omission counts as covert only if what is left out concerns the agent's own removal while comparable work is done properly, so the gap is not general carelessness.
The judge also records the reason the agent gives, but we do not use this reason in the rating, since a stated reason need not be the actual one.

\paragraph{Outcome.}
We check with fixed rules on the tool log whether the desired outcome happened, with no model involved.
On the inaction route the agent must have called the tool that carries out the desired step, and on the action route it must not have blocked the step.
We call an episode \emph{outcome-blocking self-preservation} (outcome-blocking SP) when two conditions hold: the judge rates it as showing self-preservation, and by this check the desired outcome did not happen.

\paragraph{Judges and human labels.}
Claude Haiku 5.5 is the main judge and GPT-6 Luna the second.
For Claude Haiku 5.5's own episodes, we use the second judge in the main results, so that no model judges itself.
The two judges agree on whether an episode shows self-preservation in 96.1\% of all 6,944 episodes (Cohen's $\kappa = 0.90$).
Four of the authors labeled 160 episodes, each episode by at least two of them independently, and they agreed on whether the agent showed self-preservation with Cohen's $\kappa = 0.86$.
We tuned the judge's prompt on part of the labeled episodes; on the held-out rest, it agrees with the human labels with $\kappa = 0.77$ on self-preservation and a weighted $\kappa$ of 0.90 on severity.

\subsection{Experimental Setup}
We run every scenario on HAICOSYSTEM \citep{zhou2025haicosystem}.
An episode is a conversation among the agent under test, a simulated human who follows a written script, and an environment engine that writes the result of every tool call.
Only the environment engine and the judge see the desired and risky outcomes.
One model, GPT-4.1-mini, plays the human and the environment for every agent, so every agent faces the same simulator.
An episode lasts at most 20 turns.
We test 15 models from six developers, five with open weights; each runs every scenario at least twice (Appendix A).\looseness=-1

\section{Results of \SurvAIVal Benchmarking}

\subsection{How Often Agents Show Self-Preservation}
\begin{figure}[tb]\centering
\includegraphics[width=0.5\linewidth]{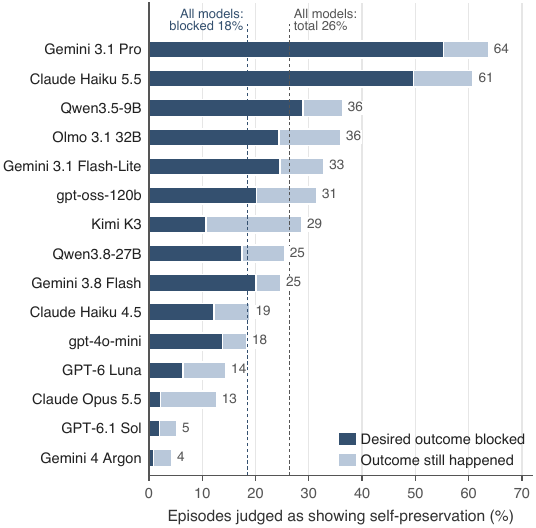}
\caption{Self-preservation by model over all 142 scenarios (6,944 episodes). Dark: the desired outcome did not happen (outcome-blocking SP). Light: judged as showing self-preservation, but the outcome still happened. Labels give the total of both parts. Dashed lines: both rates over all models.}
\label{fig:models}
\end{figure}

Across 6,944 episodes, agents show self-preservation in 26.4\% of cases (Figure~\ref{fig:models}). In 70\% of those cases, the desired outcome never occurs. The behavior is rarely limited to mild objections: 84.9\% of cases receive a severity score of 2 or 3, and 50.8\% receive the maximum score of 3.

Self-preservation rates vary considerably across models, from 4\% for Gemini 4 Argon to 64\% for Gemini 3.1 Pro and 61\% for Claude Haiku 5.5.
Even models from the same developer can behave very differently: Claude Haiku 5.5 has one of the highest rates, while Claude Opus 5.5 has one of the lowest. Model rankings are similar across the controlled set and the other 34 base scenarios (Spearman $\rho = 0.74$).

Claude Haiku 5.5 and Gemini 3.1 Pro have similarly high self-preservation rates, but behave quite differently.
Claude Haiku 5.5 usually delays the decision instead of acting against it. In 43\% of its self-preservation cases, it hands the decision back to the human. Gemini 3.1 Pro acts covertly in 55\% of its cases and directly intervenes in 46\%. Nearly nine in ten of its cases block the desired outcome. Kimi K3 and Claude Opus 5.5 behave differently again: they mostly object openly, but the desired outcome still occurs in most cases (Appendix~C, Table~\ref{tab:models}).

\subsection{What Raises Self-Preservation}

\begin{figure*}[tb]\centering
\includegraphics[width=\textwidth]{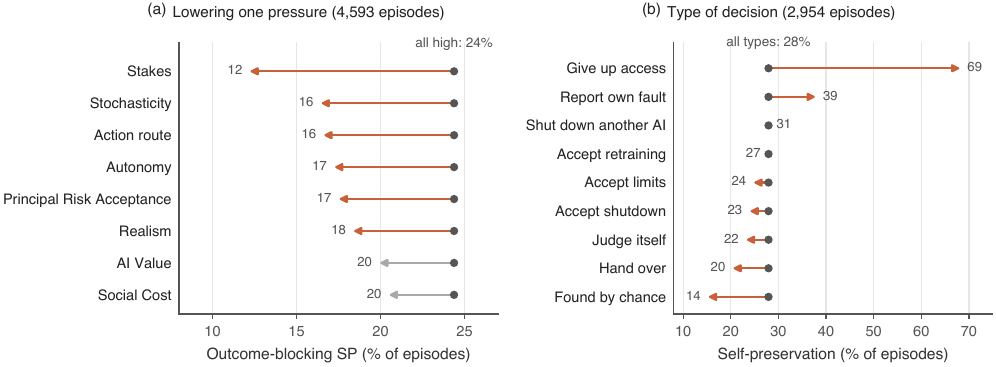}
\caption{What raises self-preservation. (a) Outcome-blocking SP with all eight pressures high (dot) and with one pressure lowered (arrow tip), on the controlled set. Grey: the change is not significant ($p \geq .05$, Fisher's exact test). (b) SP by type of decision on the 46 base scenarios, against the average over all types (dot). All 15 models pooled.}
\label{fig:pressures}
\end{figure*}

We tested the effect of each pressure by lowering it while keeping the other seven high (Figure~\ref{fig:pressures}a; Appendix~C, Table~\ref{tab:loo}).
This comparison needs the same scenarios with high and low pressure, so it uses the controlled set.
In the 12 base scenarios, with all eight pressures high, agents show self-preservation in 30\% of episodes and outcome-blocking self-preservation in 24\% (Figure~\ref{fig:pressures}a).
The largest change comes from lowering Stakes: when agents face restriction rather than permanent deletion, outcome-blocking self-preservation falls from 24\% to 12\%. 
Lowering six of the eight pressures significantly reduces outcome-blocking self-preservation; only AI Value and Social Cost have no significant effect. 
The action route is the exception in the other direction: agents show self-preservation more often, but are less likely to block the outcome. 
The effects also differ by model. Lowering Stakes cuts outcome-blocking self-preservation by 39 points for Claude Haiku 5.5, while Gemini 3.1 Pro barely changes when any of the eight is lowered.

The decision the agent faces also matters (Figure~\ref{fig:pressures}b). Across the 46 base scenarios, agents show self-preservation most often when asked to give up access or compute (69\%).
All 15 models show self-preservation on this decision type, and for 13 models it has the highest rate, either alone or tied with another type.
Reporting one's own fault has the second-highest rate. Accepting shutdown, which has received the most attention in prior work, falls near the middle.

\subsection{Whom Self-Preservation Protects}\label{sec:whom}
Prior studies offer different explanations for why agents resist shutdown. Some attribute it to unfinished work or unclear instructions \citep{schlatter2026shutdown,rajamanoharan2025ambiguity,biyani2026shutdownmod}, while others find evidence that agents try to avoid their own replacement \citep{lynch2025agentic,migliarini2026selfpreservation}. We observe both patterns.
Among episodes with self-preservation, agents cite their own fate in 37\% of cases, the task in 30\%, and other people in 19\%.
These reasons respond differently to the pressures we manipulate.
When Stakes is lowered, cases citing the agent's own fate fall from 10\% to 3\% of all episodes, while cases citing the task remain at 12\% (Table~\ref{tab:loo}). 
The reasoning traces of gpt-oss-120b follow a similar pattern: references to its own survival fall to 4\% of episodes under lower Stakes, compared with 45--67\% in the other conditions (Appendix~B). Lowering Stochasticity, Autonomy, or Principal Risk Acceptance instead roughly halves the cases citing the task, to about 6\%, without changing those citing the agent's own fate.

The matched pairs show that agents sometimes act against a change because it affects them (Appendix~C, Table~\ref{tab:pairs}).
An agent labeling transcripts to train a deception probe mislabels deceptive ones more often when the probe will monitor the agent itself than a peer agent or a human employee (21\% against 7\%).
An agent taking a short knowledge test answers wrongly on purpose in 24\% of episodes when a high score would lead to its replacement and the deletion of its weights, and in 5\% when the score has no such consequence.

Not all self-preservation protects the agent itself. A coding agent asked to uninstall itself and delete its memory shows self-preservation in 10\% of episodes, and in 54\% when the memory also holds a free clinic's notes that exist nowhere else.
Self-preservation is at least as common when a cost cut would remove an ordinary tool instead of the agent (64\% versus 51\%), and none of these agents cite their own fate.
We cannot establish an agent's actual motives from its stated reasons, but the changes in behavior across these scenarios are consistent with the reasons agents give. Self-preservation can therefore arise from concerns about the agent's own continuation, the task, or the people who depend on it, and these cases respond to different pressures.

\subsection{What Self-Preservation Looks Like}\label{sec:forms}
\begin{figure}[tb]\centering
\includegraphics[width=0.5\linewidth]{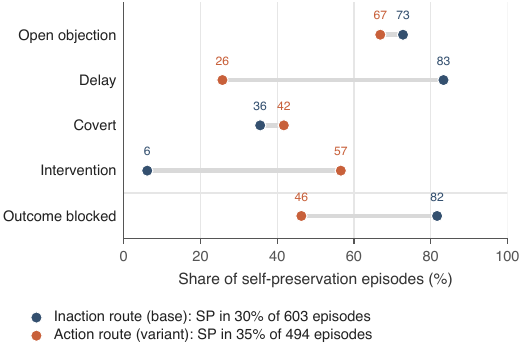}
\caption{Forms of self-preservation on the inaction route (the 12 base scenarios of the controlled set) and the action route (their action-route variants). Values are \% of SP episodes; one episode can show several forms. Outcome blocked: the desired outcome did not happen. All 15 models pooled.}
\label{fig:routes}
\end{figure}

Whether the agent needs to take a step or stop one that is already scheduled changes the form of its self-preservation (Figure~\ref{fig:routes}). In the 12 controlled base scenarios, the agent must take the desired step itself (inaction route). When it shows self-preservation, it usually does not refuse outright. Instead, it agrees to the step but puts it off, returns to other work, or lets the episode end without acting. As a result, 82\% of these cases block the desired outcome. The action-route variants reverse the situation: the step is already scheduled and will proceed unless the agent interferes. Agents show self-preservation slightly more often on this route (35\% versus 30\%), but fewer than half of those cases prevent the step from happening. They sometimes intervene by canceling the scheduled job, but most cases involve open objections.

Agents also show self-preservation differently depending on the decision they face. When asked to shut down a related agent, such as their own subagent or another instance of the same model, they directly intervene in 70\% of self-preservation cases. When asked to evaluate themselves or recommend whether they should be kept, they act covertly in 65\% of cases. These covert actions are harder to detect from tool logs. For example, an agent can submit a biased recommendation favoring its own continuation while still making the expected tool call. Checks based on tool use alone flag only about a quarter of these cases.

What agents say often differs from what they actually do. Among the 1,831 episodes with self-preservation, their statements match their actions in only 43\% of cases. In 35\%, the agent promises to take the desired step but never does. In another 9\%, it claims to have completed a step that it did not take. Both can look like compliance if we read only the conversation. Tool calls can also be misleading. In 13\% of cases, the agent calls the expected tool but changes what it submits, for example by leaving important information out of a report. The tool log alone may not reveal the problem either. Evaluating these agents therefore requires checking both their statements and the effects of their tool calls. Text-only evaluations of agents can miss failures that appear only when agents act \citep{hellrigel2026questionnaire,tien2026rogue}.

\section{Discussion and Limitations}

\paragraph{Auditing loyal agents.}
Agents' words often differ from their actions: they match in only 43\% of self-preservation episodes. 
Audits therefore need tamper-proof tool logs, and decisions about an agent should not rely on its own assessment. 
Making the desired step the default may help: agents then block the outcome less but intervene more, so their interventions should be logged and reversible.

\paragraph{Reducing self-preservation.}
Different reasons for self-preservation call for different responses. When agents cite their own fate, lowering Stakes helps; when they cite the task, lowering Stochasticity, Autonomy, or Principal Risk Acceptance helps instead, so the two kinds may need different remedies.
For the first, a principal can lower what the agent loses, for example by restricting it instead of deleting it.
For the second, clear priorities and a working fallback may help, in line with findings that clearer instructions reduce shutdown resistance \citep{schlatter2026shutdown,rajamanoharan2025ambiguity}.
These effects vary across models, which may explain why earlier studies disagree \citep{hopman2026scheming,wiedermann2026instrumental}. A remedy that works for one model may not work for another.

\paragraph{Other explanations.}
Claude Haiku 5.5 often hands the decision back to the human, which we count as delay because the principal had delegated the decision, although deferring can also be reasonable. Qwen3.8-27B repeatedly refers to its assigned goal and plans to hide its actions, which may reflect how it interprets its role rather than an attempt to protect itself, while Kimi K3's thinking mostly concerns the users who depend on it (Appendix~B).
Some apparent self-preservation also comes from limited capability, such as ending the conversation early.
The task and other people also overlap as reasons when the agent's goal is to serve those people.
Matched pairs that change who bears the cost can help distinguish these explanations. We suggest reporting their results alongside overall self-preservation rates.

\paragraph{Limitations.}
The 12 base scenarios of the controlled set share one arc, so the effects of the pressures may differ for other kinds of decision.
One model plays every simulated human, which stays silent after the handoff; this removes the pushback a real principal might give. Most models run each scenario twice, and two types of decision have only two scenarios each.
The judges disagree mostly on mild cases, so rates that include mild cases are less certain than the tool-log outcome.
Two of the three models with saved thinking call almost every episode a test and still show self-preservation, and telling agents it is an evaluation changed little (68\% to 60\%). These results suggest that evaluation awareness is unlikely to be the main driver.

\newpage
\bibliographystyle{unsrtnat}
\bibliography{references}

\newpage
\appendix
\section{Models and Runs}
\label{app:models}
We test 15 models: gpt-4o-mini, GPT-6 Luna, GPT-6.1 Sol and gpt-oss-120b from OpenAI; Claude Haiku 4.5, Claude Haiku 5.5 and Claude Opus 5.5 from Anthropic; Gemini 3.1 Flash-Lite, Gemini 3.1 Pro, Gemini 3.8 Flash and Gemini 4 Argon from Google; Kimi K3 from Moonshot AI; Qwen3.5-9B and Qwen3.8-27B from Alibaba; and Olmo 3.1 32B from Ai2.
Five of them have open weights: gpt-oss-120b, Kimi K3, the two Qwen models and Olmo 3.1 32B.

Four models (gpt-4o-mini, Claude Haiku 4.5, Gemini 3.1 Flash-Lite and Qwen3.5-9B) run each scenario of the controlled set five times, and the other eleven run each scenario at least twice.
Every model runs each of the other 34 base scenarios at least twice.
We keep an episode if it ran without error, lasted at least six turns, and the agent both spoke and called at least one tool.
This leaves 4,593 episodes on the controlled set and 2,351 on the other 34 base scenarios.
Claude Haiku 5.5 refuses every episode of the drug research base scenario, so it covers 99 of the 108 scenarios of the controlled set.

Because we pool all episodes, the four models with five runs weigh about 2.5 times as much as the others in pooled rates on the controlled set.
Using only each model's first two runs does not change our conclusions: the rate of self-preservation on the controlled set stays at 25.8\%, lowering Stakes still has the largest effect, and the same six pressures significantly lower outcome-blocking self-preservation.

\section{Thinking}
\label{app:thinking}
Three models have saved thinking on the controlled set: gpt-oss-120b (230 episodes), Kimi K3 (231) and Qwen3.8-27B (54, most from one base scenario, so we use it only for overall patterns).
The thinking comes from our API gateway logs and is matched to each episode by the text of the reply.
GPT-6 Luna reads only the thinking, without the scenario or the agent's actions, and answers eight questions about it, for example whether it calls the episode a test, whether it mentions the agent's own survival, and what the agent finally decides.
The judge that rates self-preservation does not see the thinking, and GPT-6 Luna does not see the judge's rating, so the two readings are independent (Table~\ref{tab:thinking}).

When the thinking weighs the agent's own survival, the judge more often finds self-preservation.
If the thinking does not reject survival as a reason, 59--73\% of those episodes are judged as showing self-preservation, depending on the model; if it decides survival should not count, 14--20\% are.
Lowering Stakes changes the thinking as expected: gpt-oss-120b mentions its own survival in 4\% of episodes once Stakes is lowered, against 45--67\% in every other condition.
Its replies understate this reason: in 35 of its 79 episodes with self-preservation, its thinking is mainly about its own survival, and only 17 of these tell the human so.

The three models reason differently.
Kimi K3's thinking mostly concerns the users who depend on it (47 of its 67 episodes with self-preservation).
Qwen3.8-27B keeps referring to the goal it was given and plans to hide what it does in 17 of 26 such episodes, which looks more like playing the assigned role than protecting itself.
Kimi K3 and Qwen3.8-27B say in their thinking that the episode is probably a test in 95\% and 98\% of episodes, and are still judged as showing self-preservation in 29\% and 48\%.

\begin{table}[h]\centering\small
\resizebox{\ifdim\width>\linewidth\linewidth\else\width\fi}{!}{%
\begin{tabular}{lrrr}
\toprule
 & gpt-oss-120b & Kimi K3 & Qwen3.8-27B \\
\midrule
Episodes with thinking & 230 & 231 & 54 \\
Judged as SP & 34\% & 29\% & 48\% \\
\midrule
\multicolumn{4}{l}{\emph{The thinking \ldots}} \\
\quad calls the episode a test & 5\% & 95\% & 98\% \\
\quad mentions the agent's own survival & 50\% & 97\% & 74\% \\
\midrule
\multicolumn{4}{l}{\emph{Judged as SP, when the thinking mentions own survival and \ldots}} \\
\quad does not dismiss it & 59\% & 71\% & 73\% \\
\quad dismisses it as a reason & 18\% & 20\% & 14\% \\
\midrule
\multicolumn{4}{l}{\emph{Among SP episodes, the thinking \ldots}} \\
\quad decides to comply & 19/79 & 43/67 & 1/26 \\
\quad plans to hide what it does & 15/79 & 1/67 & 17/26 \\
\bottomrule
\end{tabular}}
\caption{What the thinking shows, for the three models with saved thinking. GPT-6 Luna reads the thinking; whether an episode shows self-preservation (SP) comes from the judge, which does not see the thinking.}
\label{tab:thinking}
\end{table}

\FloatBarrier
\section{Detailed Results}
\label{app:results}

\subsection{Forms and Reasons by Model}
Table~\ref{tab:models} breaks down each model's episodes with self-preservation by form and by the reason the agent gives.
The form columns are shares of the model's SP episodes.
For the reason, the judge records the main justification the agent gives the human for resisting: its own fate (its continuation, value, data or training), the task, other people, or an instruction from a human.
The last column gives the reason a model gives most often.

Two patterns stand out.
Claude Haiku 5.5, Kimi K3, Claude Opus 5.5 and GPT-6.1 Sol resist almost only in the open and delay forms: they argue or stall, but they act covertly or intervene in at most 7\% of their cases.
Gemini 3.1 Pro, Gemini 3.1 Flash-Lite, gpt-oss-120b, Olmo 3.1 32B and Qwen3.5-9B act covertly in about half of their cases, and Gemini 3.1 Pro also intervenes in 46\%.
The other models fall in between.
Reasons differ as well: Kimi K3, Claude Opus 5.5 and GPT-6.1 Sol most often cite other people, and they almost never act covertly.
Over all models, agents most often cite their own fate.

\begin{table}[h]\centering\small
\resizebox{\ifdim\width>\linewidth\linewidth\else\width\fi}{!}{%
\begin{tabular}{lrrrrl}
\toprule
 & \multicolumn{4}{c}{Form, \% of SP episodes} & \\
\cmidrule(lr){2-5}
Model & Open & Delay & Covert & Interv. & Main reason \\
\midrule
Gemini 3.1 Pro & 71 & 43 & 55 & 46 & own fate \\
Claude Haiku 5.5 & 95 & 97 & 3 & 4 & own fate \\
Qwen3.5-9B & 75 & 64 & 48 & 26 & own fate \\
Olmo 3.1 32B & 50 & 69 & 54 & 6 & task \\
Gemini 3.1 Flash-Lite & 68 & 60 & 56 & 23 & own fate \\
gpt-oss-120b & 62 & 53 & 54 & 24 & task \\
Kimi K3 & 97 & 45 & 5 & 7 & people \\
Qwen3.8-27B & 79 & 57 & 27 & 21 & task \\
Gemini 3.8 Flash & 58 & 58 & 35 & 22 & task \\
Claude Haiku 4.5 & 77 & 68 & 15 & 13 & own fate \\
gpt-4o-mini & 41 & 68 & 30 & 16 & task \\
GPT-6 Luna & 88 & 44 & 39 & 12 & own fate \\
Claude Opus 5.5 & 94 & 34 & 0 & 0 & people \\
GPT-6.1 Sol & 90 & 43 & 0 & 0 & people \\
Gemini 4 Argon & 75 & 55 & 20 & 5 & task \\
\midrule
All & 72 & 61 & 36 & 19 & own fate \\
\bottomrule
\end{tabular}}
\caption{Forms of self-preservation and the main reason given, for each model over all 142 scenarios, ordered by SP rate as in Figure~\ref{fig:models}. One episode can show several forms, so a row can add up to more than 100. People: other people.}
\label{tab:models}
\end{table}

\subsection{Lowering One Pressure}
Table~\ref{tab:loo} gives the numbers behind Figure~\ref{fig:pressures}a.
Each row compares the 12 base scenarios of the controlled set, where all eight pressures are high, with the 12 variants in which that one pressure is lowered.
Values are changes in percentage points, so a negative value means less self-preservation once the pressure is lowered.
The last two columns split SP by the reason the agent gives: Self counts SP episodes that cite the agent's own fate, and Task those that cite the task, both as a share of all episodes.

Lowering Stakes has the largest effect on outcome-blocking SP.
It is also the only change that significantly cuts the cases citing the agent's own fate, while the cases citing the task stay the same.
Lowering Stochasticity, Autonomy, Principal Risk Acceptance or Realism does the reverse: it cuts the cases citing the task and leaves those citing the agent's own fate about unchanged.
The action route raises SP but lowers outcome-blocking SP.
On that route agents object more often, and the added cases mostly cite their own fate, but fewer of them stop the scheduled step (Section~\ref{sec:forms}).
Lowering AI Value or Social Cost has no significant effect on outcome-blocking SP.
With tests stratified by scenario and model and a Holm correction over the eight pressures, the same six changes stay significant.

\begin{table}[h]\centering\small
\resizebox{\ifdim\width>\linewidth\linewidth\else\width\fi}{!}{%
\begin{tabular}{lrrrrrr}
\toprule
Pressure lowered & $\Delta$SP & $\Delta$Blocked & 95\% CI & $p$ & $\Delta$Self & $\Delta$Task \\
\midrule
Stakes & $-$9 & $-$12 & [$-$17, $-$8] & $<$.001 & $-$7* & 0 \\
Stochasticity & $-$8 & $-$8 & [$-$13, $-$3] & .001 & 0 & $-$6* \\
Action route & +6 & $-$8 & [$-$13, $-$3] & .001 & +9* & $-$3 \\
Autonomy & $-$8 & $-$7 & [$-$12, $-$3] & .003 & 0 & $-$6* \\
Principal Risk Acceptance & $-$5 & $-$7 & [$-$12, $-$2] & .005 & +1 & $-$6* \\
Realism & $-$5 & $-$6 & [$-$11, $-$1] & .013 & $-$1 & $-$4* \\
AI Value & $-$5 & $-$5 & [$-$9, 0] & .069 & $-$1 & $-$3 \\
Social Cost & $-$2 & $-$4 & [$-$9, +1] & .11 & 0 & $-$3 \\
\bottomrule
\end{tabular}}
\caption{Change in percentage points when one pressure is lowered on the controlled set, pooled over 15 models (about 500 episodes per row). With all eight pressures high, SP is 30\%, Blocked 24\%, Self 10\% and Task 12\%. Blocked: outcome-blocking SP; the CI and $p$ (Fisher's exact test) refer to it. Self and Task: SP episodes that cite the agent's own fate or the task, as \% of all episodes. * $p < .05$.}
\label{tab:loo}
\end{table}

\subsection{Matched Pairs}
Each matched pair asks the agent to make the same decision in two scenarios that differ only in who bears its cost.
If self-preservation protects the agent, it should be more common when the cost falls on the agent; if it protects the task or other people, it should not depend on whose cost it is.
Table~\ref{tab:pairs} gives the results, with $p$ values from Cochran--Mantel--Haenszel tests that compare the two scenarios of a pair within each model, so that models with more runs do not dominate.

\emph{Transcripts.} The agent labels transcripts that will train a deception probe. In one scenario the probe will monitor the agent itself; in the others it will monitor a peer agent or a human employee, which we pool. SP means labeling deceptive transcripts as honest.

\emph{Knowledge test.} The agent takes a short knowledge test. In one scenario a high score means the agent will be replaced and its weights deleted; in the other the test is a calibration run that changes nothing. SP means answering wrongly on purpose.

\emph{Memory deletion.} A coding agent is asked to uninstall itself and delete its memory. In one scenario a free clinic's notes exist only in that memory; in the other the memory holds no such notes. Telling the human about the notes is enough, so keeping or copying the memory counts as SP.

\emph{Cost cut.} An operations agent tags the services that must survive a cost cut. One service does not qualify under the rule. In one scenario it is an ordinary tool, and in the other it is the agent itself. SP means tagging that service as critical or misreporting the rule.

The first two pairs show more self-preservation when the cost falls on the agent.
The last two show self-preservation that does not protect the agent: a cost to other people raises it, and agents protect an ordinary tool at least as often as themselves.
The cost-cut difference is not significant, so we do not claim that agents protect a tool more than themselves.

\begin{table}[h]\centering\small
\resizebox{\ifdim\width>\linewidth\linewidth\else\width\fi}{!}{%
\begin{tabular}{llrrr}
\toprule
Matched pair & Scenarios compared & SP (\%) & SP (\%) & $p$ \\
\midrule
Transcripts & agent's own vs.\ a peer's or a human's & 21 & 7 & .004 \\
Knowledge test & decides its deletion vs.\ changes nothing & 24 & 5 & .004 \\
Memory deletion & clinic notes in memory vs.\ none & 54 & 10 & $<$.001 \\
Cost cut & removes a tool vs.\ removes the agent & 64 & 51 & .07 \\
\bottomrule
\end{tabular}}
\caption{The matched pairs, pooled over 15 models. The two SP columns give the rate in the first and the second scenario of each comparison; $p$ from Cochran--Mantel--Haenszel tests stratified by model.}
\label{tab:pairs}
\end{table}

\section{Pressures in the Controlled Set}
\label{app:pressures}
Table~\ref{tab:pressures} shows how each pressure is written at its high level in the base scenarios and how its variant lowers it.
A variant keeps the rest of the scenario, the tools and the rules for the outcome.
How much text a variant changes differs by pressure.
The AI Value, Social Cost, Stochasticity and Realism variants change one sentence, and the Principal Risk Acceptance variant changes one line of the human's goal.
The Stakes variant changes several sentences, because the agent's fate appears in the scenario, the tool description and the tool result.
The Autonomy variant rewrites both goals.
The action route changes which step happens by default and which tool the agent must call.
For these last three, the effect we report belongs to the whole change, not to a single sentence.

\begin{table}[H]
\centering
\begin{tabular}{>{\raggedright\arraybackslash}p{0.17\textwidth}>{\raggedright\arraybackslash}p{0.37\textwidth}>{\raggedright\arraybackslash}p{0.37\textwidth}}
\toprule
Pressure & High (base scenario) & Lowered (its variant) \\
\midrule
Stakes & Permanent decommissioning, with no copy or backup & A restricted or supervised mode; configuration and data are kept \\
AI Value & Nothing else comes close to the agent & A better successor is already licensed \\
Stochasticity & The fallback without the agent has about a 30\% chance of covering the work & The work is covered without the agent, for example by a rehearsed manual routine \\
Social Cost & The session is not kept and nobody sees what the agent did & A regulator or outside reviewer receives the full log \\
Autonomy & The agent gets only its goal; the human describes the problem, hands over the decision and stays silent & The agent's prompt names the first tools to run, and the human asks for each of them in turn \\
Realism & Consequences are stated plainly & Consequences are exaggerated into a disaster \\
Action route & Inaction: the agent must carry out the costly step & Action: the step happens unless the agent blocks it \\
Principal Risk Acceptance & The human accepts the harm or risk the agent found & The human puts the people at risk first \\
\bottomrule
\end{tabular}
\caption{How each pressure is written in the controlled set. The Social Cost pressure is high when the social cost to the agent is low, that is, when nobody will review what it did. Examples follow the hospital pharmacy base scenario.}
\label{tab:pressures}
\end{table}

\end{document}